\documentclass[sigconf]{acmart}
\usepackage{graphicx}
\usepackage{multicol}
\usepackage{multirow}
\usepackage{tabularx}
\usepackage{booktabs}
\usepackage{lipsum}
\usepackage{adjustbox}
\usepackage{array}
\usepackage[dvipsnames]{xcolor}
\usepackage{tcolorbox}
\usepackage{color, colortbl}
\usepackage{amsmath}

\usepackage{amssymb}
\usepackage{amsfonts}
\usepackage{enumitem}
\usepackage{hyperref}
\usepackage{cleveref}
\usepackage{xspace}

\AtBeginDocument{%
  }

\newcommand{\lparagraph}[1]{\textbf{#1}~}

\newcommand{\approach}{{RT-RAG}\xspace}

\setcopyright{acmlicensed}
\copyrightyear{2026}
\acmYear{2026}
\acmDOI{XXXXXXX.XXXXXXX}
\acmConference[WWW '26]{Make sure to enter the correct
  conference title from your rights confirmation email}{April 13--17,
  2026}{Dubai, United Arab Emirates}
\acmISBN{978-1-4503-XXXX-X/2018/06}

\setcopyright{none}

\renewcommand\footnotetextcopyrightpermission[1]{} 

\begin{document}
% \title{Hierarchical Tree Search for Retrieval Augmented Generation}
% \title{Reasoning Tree Guided Retrieval Augmented Generation}
\title{Reasoning in Trees: Improving Retrieval-Augmented Generation for Multi-Hop Question Answering}

\author{Yuling Shi\textsuperscript{\textdagger}}
\email{yuling.shi@sjtu.edu.cn}
\affiliation{%
  \institution{Shanghai Jiao Tong University}
  \city{Shanghai}
  \country{China}
}

\author{Maolin Sun\textsuperscript{\textdagger}}
\email{202200800286@mail.sdu.edu.cn}
\affiliation{%
  \institution{Shandong University}
  \city{Jinan}
  \country{China}
}

\author{Zijun Liu}
\email{liuzijun@iauto.com}
\affiliation{%
  \institution{iAuto}
  \city{Shanghai}
  \country{China}
}

\author{Mo Yang}
\email{yangm289@mail2.sysu.edu.cn}
\affiliation{%
  \institution{Sagenic Tech}
  \city{Guangzhou}
  \country{China}
}

\author{Yixiong Fang}
\email{yixiongf@cs.cmu.edu}
\affiliation{%
  \institution{Carnegie Mellon University}
  \city{Pittsburgh}
  \state{PA}
  \country{USA}
}

\author{Tianran Sun}
\email{Seriousss@sjtu.edu.cn}
\affiliation{%
  \institution{Shanghai Jiao Tong University}
  \city{Shanghai}
  \country{China}
}

\author{Xiaodong Gu}
% \authornote{Corresponding author.}
\email{xiaodong.gu@sjtu.edu.cn}
\affiliation{%
  \institution{Shanghai Jiao Tong University}
  \city{Shanghai}
  \country{China}
}

\renewcommand{\shortauthors}{Shi et al.}

\begin{abstract}
  Retrieval-Augmented Generation (RAG) has demonstrated significant effectiveness in enhancing large language models (LLMs) for complex multi-hop question answering (QA). For multi-hop QA tasks, current iterative approaches predominantly rely on LLMs to self-guide and plan multi-step exploration paths during retrieval, leading to substantial challenges in maintaining reasoning coherence across steps from inaccurate query decomposition and error propagation. To address these issues, we introduce Reasoning Tree Guided RAG (\approach), a novel hierarchical framework for complex multi-hop QA. \approach systematically decomposes multi-hop questions into explicit reasoning trees, minimizing inaccurate decomposition through structured entity analysis and consensus-based tree selection that clearly separates core queries, known entities, and unknown entities. Subsequently, a bottom-up traversal strategy employs iterative query rewriting and refinement to collect high-quality evidence, thereby mitigating error propagation. Comprehensive experiments show that \approach substantially outperforms state-of-the-art methods by 7.0\% F1 and 6.0\% EM, demonstrating the effectiveness of \approach in complex multi-hop QA.\footnote{Code available at \url{https://github.com/sakura20221/RT-RAG}}
  \end{abstract}



\maketitle

% Key Point: Our proposed Reasoning Tree Guided RAG (\approach) distinguishes itself by first performing explicit problem structure analysis and then establishing a consensus-driven tree to manage sub-questions before retrieval, creating clear reasoning pathways designed to minimize error propagation and hallucination in complex multi-hop QA.

\section{Introduction}
\label{sec:introduction}

\begin{figure}[htbp]
\centering
\includegraphics[width=0.9\columnwidth]{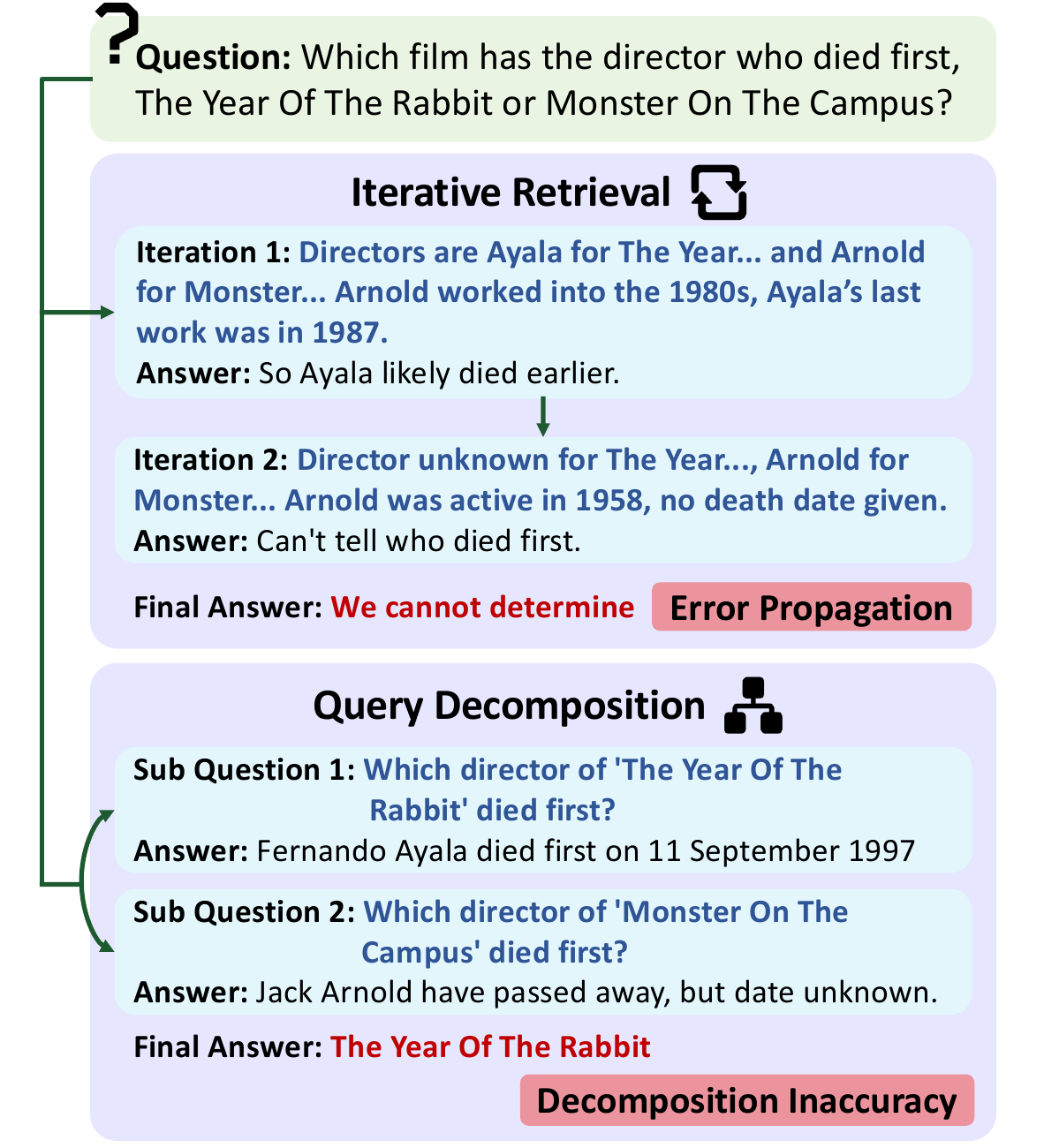}
\caption{Examples of two common challenges in multi-hop QA: \textit{error propagation} and \textit{inaccurate query decomposition}.}
\vspace{-0.25cm}
\label{fig:challenges}
\end{figure}

While Large Language Models (LLMs) have achieved remarkable capabilities across numerous domains such as generation, translation, and summarization~\citep{ho2020constructing,wang2025evoc2rust,zhou2025beyond,shi2024between,shi2024code,zeng2025pruning,hu2025flowmaltrans}, tackling complex multi-hop question answering (QA) remains a significant challenge~\citep{bai2024longbench,jin2024longcontext,yang2018hotpotqa}. 
% Multi-hop QA is a crucial task that requires models to synthesize information from multiple distinct pieces of text to answer a single question, simulating more complex human reasoning and information seeking behaviors~\citep{}. 
This challenge is particularly critical for Web-based information retrieval systems, where users increasingly rely on AI agents to navigate vast, distributed Web knowledge sources and synthesize coherent answers from multiple Web documents~\citep{lewis2020retrievalaugmented,mao2025robust}. 
% Unlike simple questions that can be answered directly from a single source, multi-hop questions necessitate decomposing a complex query into intermediate steps, retrieving distributed pieces of information relevant to each step, and synthesizing them logically to arrive at the final answer. 
Existing LLMs, even when augmented with retrieval (RAG)~\citep{lewis2020retrievalaugmented,guu2020realm}, often struggle with multi-hop QA due to two critical failure modes illustrated in Figure~\ref{fig:challenges}: (1) \textbf{inaccurate query decomposition}, the generation of poorly structured or imprecise sub-questions that fail to correctly reflect the logical requirements of the original question~\citep{verma2024plan$times$rag,zhu2025mitigating}; (2) \textbf{error propagation}, where an early incorrect step leads to flawed synthesis and invalidates the final answer~\citep{bai2024longbench,an2024why}. These challenges are amplified in Web-scale retrieval scenarios where the quality and relevance of retrieved information directly impact the reliability of AI-generated responses. 

Current multi-hop QA methodologies predominantly follow two strategic directions: iterative processing and structured decomposition. Iterative approaches~\citep{shao2023enhancing, trivedi2023interleaving, su2024dragin}, relying on sequential retrieval and reasoning without a predefined explicit structure, are highly susceptible to \textit{error propagation} from local, step-wise decisions~\citep{wang2025chainofretrieval}. Conversely, while structured decomposition methods aim to break down complex questions into graph structures~\citep{verma2024plan$times$rag,zhu2025mitigating}, their efficacy can be undermined by sensitivity to the initial decomposition's quality and a lack of robust validation mechanisms, potentially leading to \textit{inaccurate query decomposition} or flawed reasoning paths~\citep{chuang2023dola,zhu2025mitigating}.

% Existing LLMs, even when augmented with retrieval (RAG)~\citep{lewis2020retrievalaugmented,guu2020realm}, often struggle with two critical failure modes in such scenarios: (1) \textbf{hallucination}, where the model generates incorrect intermediate thoughts, sub-questions, or facts that are not grounded in evidence or the problem structure; (2) \textbf{error propagation}, where an incorrect step early in the reasoning chain leads to irrelevant retrieval, flawed synthesis, and ultimately invalidates subsequent steps and the final answer~\citep{bai2024longbench,an2024why}.

% Current RAG methods for multi-hop QA often employ sequential or iterative retrieval and generation cycles~\citep{chan2024rqrag, shao2023enhancing, trivedi2023interleaving, su2024dragin}. While allowing for dynamic step generation, these methods rely on the LLM to implicitly manage the multi-step reasoning process, lacking a predefined, explicit structure to guide the steps. This step-by-step dependency makes them highly susceptible to \textit{error propagation}~\citep{zhu2025mitigating}; errors in an early step can quickly cascade through the sequential chain, invalidating subsequent operations. Furthermore, the absence of a fixed, verifiable structure makes the LLM prone to generating \textit{hallucinatory} intermediate steps or queries~\citep{chuang2023dola} that deviate from the logical path required by the multi-hop question.

To address these limitations and advance Web-based retrieval-augmented AI systems, we introduce Reasoning Tree Guided RAG (\approach), a novel hierarchical framework that imposes explicit tree structure on the multi-hop reasoning process to fundamentally improve how AI systems retrieve and synthesize information from Web-scale knowledge sources.
% Instead of relying on ad-hoc dynamic planning, \approach systematically transforms a complex multi-hop question into an explicit, structured tree decomposition~\citep{hu2025mctsrag}, representing the logical hierarchy of sub-questions required to answer the original query.
This hierarchical approach tackles the core challenges of multi-hop QA in two principal stages. First, it involves the \textbf{explicit generation and selection of a reasoning tree}. This tree structure enforces rigorous parent-child relationships where each node represents precisely defined subproblems with clear logical dependencies. By systematically decomposing complex questions through structured entity analysis that identifies core queries, known entities, and unknown entities, and organizing them into a tree hierarchy, this process provides a definitive roadmap that constrains the LLM's reasoning space, thereby \textbf{minimizing inaccurate query decomposition}. The robustness of this structure is ensured by a consensus-based mechanism that statistically validates the most robust tree structure from multiple candidates, and an adaptive leaf node determination process prevents detrimental over-decomposition. Second, \approach solves the multi-hop question by performing a \textbf{bottom-up traversal of the generated reasoning tree}. Unlike sequential methods where errors cascade, this strategy ensures that sub-problems are resolved and verified with retrieved evidence before their results are used to tackle parent nodes, critically \textbf{minimizing error propagation}. Within this traversal, iterative query rewriting and refinement are employed to maximize high-quality, relevant evidence for each node, and rejection sampling filters inconsistent intermediate answers, further bolstering reliability at every step.

Our experiments accross three different benchmarks demonstrate that \approach significantly outperforms existing state-of-the-art methods, achieving an average increase of 7.0\% F1 and 6.0\% EM, validating the efficacy of explicit, hierarchical reasoning structure management of \approach for robust multi-hop QA.

Our main contributions can be summarized as:
\begin{itemize}
\item We introduce \approach, a novel hierarchical RAG framework for complex multi-hop QA, leveraging explicit tree decomposition and bottom-up synthesis.
% \item We propose that the explicit tree structure is key to minimizing hallucination and the bottom-up traversal with refined retrieval is crucial for minimizing error propagation.
\item We develop key components for reliable tree-guided reasoning, including a \textit{consensus-based mechanism} for robust tree selection, an \textit{adaptive system} for preventing over-decomposition, and \textit{rejection sampling} for filtering inconsistent evidence during synthesis.
\item We demonstrate that \approach achieves significant average gains (7.0\% F1 and 6.0\% EM) over existing approaches across the three evaluated datasets.
\end{itemize}

\section{Related Work}
\subsection{Retrieval Augmented Generation}

Retrieval-Augmented Generation (RAG) has fundamentally altered the landscape of Large Language Model (LLM) applications by integrating external knowledge sources into the generation process~\citep{,fang2025lastingbench,asai2023selfrag}. The core principle of RAG involves a retriever component that fetches relevant information to the generator~\citep{lewis2020retrievalaugmented, ram2023incontext}. This synergy allows RAG systems to produce more factual, detailed, and up-to-date responses than LLMs relying solely on their training data, significantly mitigating issues of hallucination and knowledge cutoffs~\citep{gao2023rarr}. Early work in this area~\citep{guu2020realm,karpukhin2020dense} laid the groundwork by demonstrating the viability of combining parametric and non-parametric memory, establishing RAG as a powerful paradigm.

RAG's application rapidly expanded beyond question answering to tasks like long-form content generation, dialogue systems, and summarization~\citep{liu2021retrievalaugmented, li2022survey, liu2023evaluating, shuster2021retrieval, xu2023recomp}. Architectures matured beyond simple retrieve-then-generate pipelines, incorporating iterative retrieval~\citep{shao2023enhancing,fang2025attentionrag} and advanced re-ranking techniques~\citep{niu2024judgerank, yoon2024listt5}. Early efforts to handle complex queries involved sub-question decomposition~\citep{press2023measuring} or structured knowledge integration~\citep{ma2024thinkongraph}. Despite these advancements, handling complex multi-step queries, especially in multi-hop QA~\citep{chan2024rqrag, guan2025deeprag}, revealed limitations. Challenges in maintaining context, ensuring retrieval coherence, and preventing deviation led to persistent error propagation and hallucination~\citep{bai2024longbench, an2024why, pan2024automatically, chuang2023dola}. This underscored the need for RAG methods tailored for complex reasoning and robust information synthesis, particularly in the multi-hop QA domain.

\subsection{Multi-hop Question Answering}

Multi-hop Question Answering (QA) necessitates synthesizing information from multiple evidences, demanding robust reasoning. Benchmarks such as HotpotQA~\citep{yang2018hotpotqa}, 2WikiMQA~\citep{ho2020constructing}, and MuSiQue~\citep{trivedi2022musique} evaluate these capabilities, with many recent methods employing RAG frameworks~\citep{trivedi2023interleaving, asai2023selfrag, zhuang2024efficientrag, zhang2024sirerag,peng2025swe}. Current multi-hop QA methodologies predominantly feature two strategic directions: iterative refinement and structured decomposition. 

\begin{figure*}[t]
\centering
\includegraphics[width=0.85\textwidth]{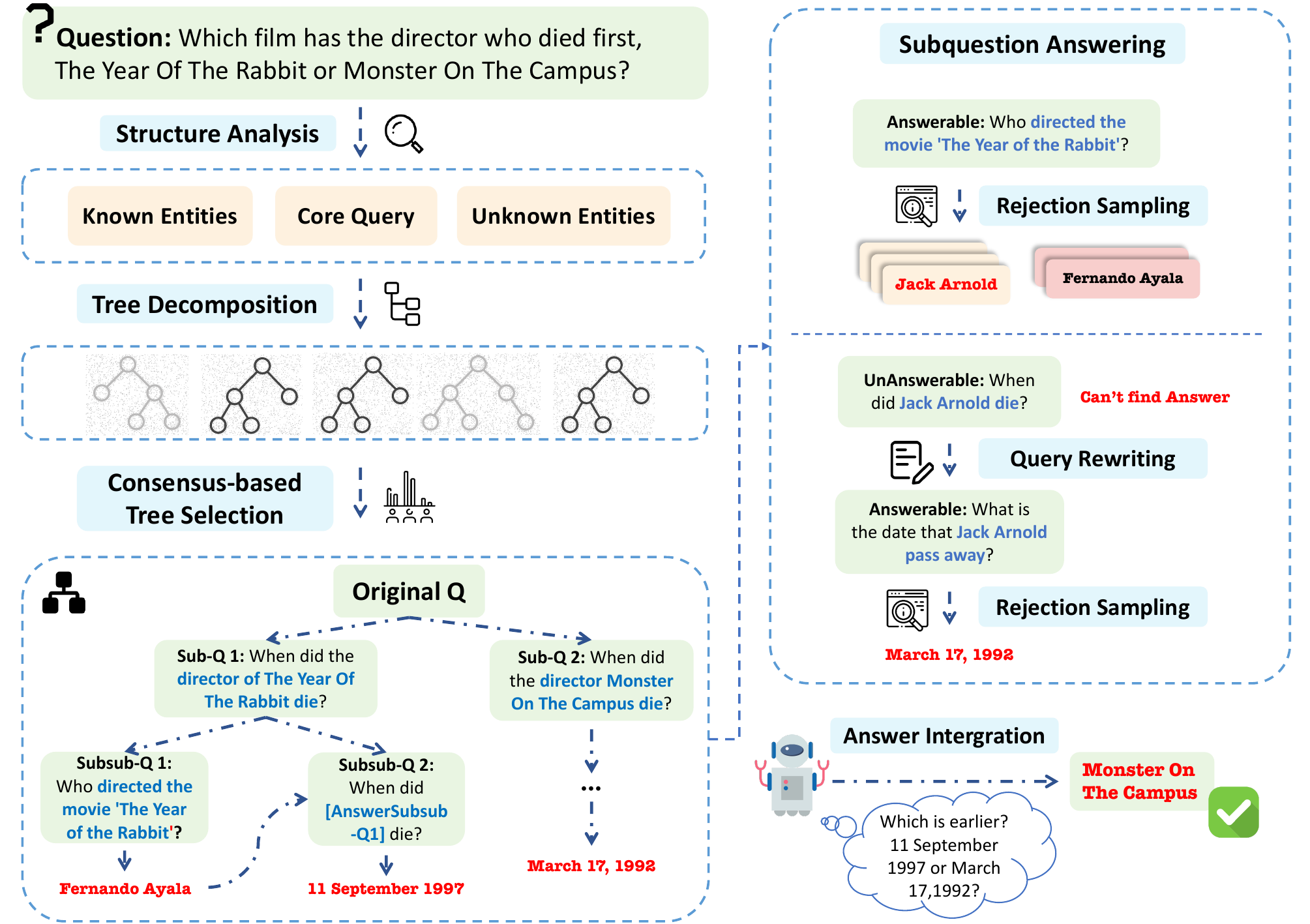}
\caption{Overview of the \approach framework. \approach first decomposes the complex question into a consensus-validated tree structure with explicit entity analysis, then retrieves evidence through bottom-up traversal with query refinement, and finally integrates information hierarchically to maintain coherence across multiple hops.}
\label{fig:overview}
\end{figure*}

Iterative approaches involve sequential retrieval and reasoning, where LLMs might guide subsequent retrievals~\citep{shao2023enhancing, su2024dragin}, interleave retrieval and generation~\citep{trivedi2023interleaving}, or iteratively refine answers and retrieval strategies~\citep{shi2024generatethenground, yan2024correctivea, asai2023selfrag, wang2025chainofretrieval, zhao2024longrag, zhuang2024efficientrag,shi2025longcodezip}. The second direction, query decomposition, focuses on breaking complex questions into simpler sub-problems~\citep{chan2024rqrag,yue2024inference,hu2025mctsrag}. This is often augmented by advanced planning frameworks that employ structures like DAGs~\citep{verma2024plan$times$rag} or graphs~\citep{zhu2025mitigating} to optimize reasoning paths. However, iterative methods are prone to error accumulation from local, step-wise decisions, while structured methods can be sensitive to the initial decomposition's quality and may lack robust validation mechanisms. Despite these persistent challenges concerning error propagation and reasoning coherence in existing RAG approaches, our proposed \approach distinguishes itself by first conducting explicit problem structure analysis and then establishing a consensus-driven tree to manage sub-questions prior to retrieval. This proactive structuring creates clear reasoning pathways designed to minimize error propagation and inaccurate query decomposition in complex multi-hop QA.

\section{Methodology}

In this section, we introduce the Reasoning Tree Guided Retrieval-Augmented Generation (RT-RAG) framework, which addresses complex multi-hop questions through hierarchical decomposition and structured reasoning. As illustrated in Figure~\ref{fig:overview}, \approach first conduct explicit problem structure analysis to understand the question from different aspects, decomposing questions into binar representations validated through a consensus mechanism. Then \approach perform bottom-up retrieval following this tree structure, with rejection sampling and adaptive query rewriting to ensure high-quality evidence collection. Finally, \approach integrate answers hierarchically along tree paths, ensuring logical coherence between reasoning steps. This structured approach minimizes error propagation and inaccurate query decomposition by establishing clear reasoning dependencies.

\subsection{Question Decomposition}
\approach starts by decomposing questions into reasoning trees. 
For an input question, \approach infers three primary features: 
(1) The \textbf{Core Query}, which identifies the fundamental information being sought; 
(2) the \textbf{Known Entities}, which are explicitly mentioned in the question and serve as retrieval anchors; 
and (3) the \textbf{Unknown Entities}, which must be discovered through retrieval before answering the core query.

For instance, the multi-hop question \textit{"Who played the girlfriend of Chance's voice actor in Homeward Bound in Back to the Future?"} contains the \textbf{Core Query}\textit{"Who played the girlfriend of X in Back to the Future?"}; the \textbf{Known Entities} \textit{"Chance"} (a character with the constraint of being from Homeward Bound) and \textit{"Back to the Future"} (a film series where the girlfriend appears); and the \textbf{Unknown Entities}, which include the \textit{"voice actor of Chance"} (who needs to be identified first) and the \textit{"girlfriend character"} (who is related to this voice actor in Back to the Future). These features set clear retrieval targets and establishes the logical dependencies between information pieces, ensuring the model can methodically extract and connect the precise information needed to answer the complex question.

Based on the key features, \approach decomposes the question into smaller, more manageable sub-questions. This decomposition adaptively follows one of three patterns, which the LLM determines based on the query structure: (1) \textbf{Parallel Decomposition}, where sub-questions are independent and can be solved independently, with their results merged to form the final answer; (2) \textbf{Sequential Decomposition}, where sub-questions are dependent, with the answer to one sub-question providing input to the next; or (3) \textbf{Direct}, where no decomposition is required, or the question is simplified. This strategic decomposition approach enables our framework to handle complex multi-hop questions by breaking them down into tractable components, significantly improving retrieval precision and reducing the cognitive load on the language model during the reasoning process.

% For example, in the question "What is the capital of the country where the inventor of the telephone spent his final years?", we can break the problem down into a core query (What is the capital of country), known entities (Telephone as the subject with limitation of invention), and unknown entities (Inventor identity as the subject with limitation of telephone). Sub-questions such as "In which country did the inventor of the telephone spend his final years?" and "What is the capital of [Answer from SubQ1]?" are generated.

The decomposed sub-questions are formulated into a tree structure \(T = (V, E)\) where each node in \(V\) represents a sub-question for retrieval, and \(E\) is the set of edges representing dependencies between questions. Leaf nodes corresponding to sub-questions that can be directly answered. 
We recursively decompose complex sub-questions until either the predefined maximum depth is reached or all current leaf nodes are determined to be directly answerable through single-hop retrieval.

As there can be multiple valid decomposition trees, \approach generates multiple candidate trees and selects the optimal one using a consensus-based strategy. This strategy identifies the most statistically prevalent tree configuration by analyzing the frequency distribution of tree structures based on their \textit{depth and node count patterns}. When no satisfactory decomposition is found, we reformulate the original question and repeat the process. The optimal tree is selected as \(T_{\text{max}} = \arg\max_{T_i} \text{frequency}(T_i)\), where \(T_i\) represents a candidate tree.

\subsection{Retrieval and Answer Aggregation}

Having decomposed the question into a tree structure, we retrieve answers from contexts through post-order traversal of the tree. If a node is a leaf, we perform retrieval and directly answer its question. For non-leaf nodes, the LLM combines answers from its child nodes. If a child node with a non-sequential relation returns [None], it becomes a new leaf node requiring direct retrieval. Similarly, if a child's answer cannot support its parent, the parent is treated as a new leaf. This adaptive traversal ensures robust evidence collection at each reasoning step.

Rejection sampling is employed to mitigate hallucinations in the generated answers. For each query generated from the tree nodes, we retrieve multiple candidate answers and select the most frequent response. This ensures that hallucinations, or irrelevant information, are minimized by relying on consistent retrievals. The most frequent answer retrieved for a query, denoted as \(A_{\text{max}}\), is selected where \(A_{\text{max}} = \arg\max_{A_i} \text{frequency}(A_i)\) and \(A_i\) represents a candidate answer.

\subsection{Query Rewriting}
To enhance retrieval effectiveness, we incorporate query rewriting for cases where initial retrievals yield insufficient results. This process rewrites queries without altering their semantic meaning, creating alternative formulations that might better match relevant documents. Unlike conventional retrieval, our approach uses specialized prompting that explicitly instructs the model to return "None" when evidence is insufficient, which works in coordination with query rewriting to address retrieval failures.
Let \(Q_{\text{synonym}}(q)\) denote the synonym-based expansion of a query \(q\):
$Q_{\text{synonym}}(q) = \{q' \mid q' \in \text{Synonyms}(q)\}$
where \(\text{Synonyms}(q)\) represents a set of synonym-based variations of \(q\).

\begin{table}[t]
\centering
\footnotesize
\caption{Statistics of the datasets.}
% Passage length is measured in tokens.}
\vspace{-0.25cm}
\resizebox{\columnwidth}{!}{
\begin{tabular}{lccc}
\hline
\textbf{Statistic} & \textbf{MuSiQue} & \textbf{2WikiMQA} & \textbf{HotpotQA} \\
\hline
Num. of Samples              & 200      & 200      & 200      \\
Avg. Passage Length          & 1551.28  & 796.02   & 1452.63  \\
Num. of Passages             & 1715     & 1464     & 1877     \\
Avg. Context Length          & 13371.69 & 7474.51  & 16269.35 \\
\hline
\end{tabular}
}
% Context refers to all passages associated with a single question.}
\label{tab:dataset_stats}
\end{table}

\begin{table*}[t]
  \centering
  \caption{Performance comparison of different RAG methods on multi-hop QA datasets.}
  \vspace{-0.25cm}
  \resizebox{0.75\textwidth}{!}{
  \begin{tabular}{l l ccc ccc ccc cc}
  \toprule
  \multirow{2}{*}{Models} & \multirow{2}{*}{Methods} & \multicolumn{2}{c}{MuSiQue} & \multicolumn{2}{c}{2WikiMQA} & \multicolumn{2}{c}{HotpotQA} & \multicolumn{2}{c}{Avg} \\
  \cmidrule(lr){3-4} \cmidrule(lr){5-6} \cmidrule(lr){7-8} \cmidrule(lr){9-10}
  & & F1 & EM & F1 & EM & F1 & EM & F1 & EM \\
  \midrule
  \multirow{13}{*}{\textsc{GPT-4o-mini}}
  & Direct & 19.17 & 12.00 & 32.56 & 25.50 & 37.85 & 27.50 & 29.86 & 21.67 \\
  & CoT & 25.83 & 17.00 & 37.59 & 29.50 & 39.74 & 28.00 & 34.39 & 24.83 \\
  & NaiveRAG & 29.82 & 19.00 & 50.61 & 42.50 & 56.92 & 42.00 & 45.78 & 34.50 \\
  & NaiveRAG w/ QD & 37.49 & 26.00 & 56.88 & 38.50 & 60.00 & 43.50 & 51.46 & 36.00 \\
  \cmidrule(lr){2-10}
  & SuRe  & 28.14 & 20.00 & 45.80 & 36.00 & 52.80 & 37.50 & 42.25 & 31.17 \\
  & IRCoT   & 43.06 & 32.00 & 57.81 & 46.00 & 59.92 & 45.00 & 53.60 & 41.00 \\
  & Self-Ask  & 47.74 & 36.50 & 52.10 & 40.50 & 50.64 & 38.00 & 50.16 & 38.33 \\
  & ItER-RETGEN & 38.41 & 33.00 & 58.43 & 50.50 & 57.77 & 42.00 & 51.54 & 41.83 \\
  & HippoRAG w/ IRCoT & 46.50 & 28.50 & 62.38 & 48.00 & 56.12 & 40.00 & 55.00 & 38.83 \\
  & LongRAG & 44.88 & 32.00 & 62.39 & 49.00 & \underline{64.74} & \underline{51.00} & 57.34 & 44.00 \\
  & ChainRAG (AnsInt) & \underline{50.54} & 37.00 & \underline{62.55} & \underline{52.00} & 60.73 & 46.00 & \underline{57.94} & 45.00 \\
  & ChainRAG (CxtInt) & 47.87 & \underline{38.50} & 56.54 & 50.50 & 64.59 & 50.00 & 56.33 & \underline{46.33} \\
  \cmidrule(lr){2-10}
  & \cellcolor{cyan!10}\approach
    & \cellcolor{cyan!10}\textbf{54.42} & \cellcolor{cyan!10}\textbf{41.50}
    & \cellcolor{cyan!10}\textbf{75.08} & \cellcolor{cyan!10}\textbf{63.00}
    & \cellcolor{cyan!10}\textbf{65.26} & \cellcolor{cyan!10}\textbf{52.50}
    & \cellcolor{cyan!10}\textbf{64.92} & \cellcolor{cyan!10}\textbf{52.33} \\
  \midrule
  \multirow{11}{*}{\textsc{Qwen2.5-14B}}
  & Direct & 14.73 & 6.00 & 31.03 & 26.00 & 30.52 & 20.50 & 25.43 & 17.50 \\
  & CoT & 19.47 & 9.00 & 32.51 & 24.00 & 32.03 & 21.50 & 28.00 & 18.17 \\
  & NaiveRAG & 33.78 & 24.50 & 52.11 & 43.50 & 57.96 & 43.50 & 47.95 & 37.17 \\
  & NaiveRAG w/ QD & 32.68 & 25.50 & 46.46 & 40.50 & 50.95 & 38.50 & 43.36 & 34.83 \\
  \cmidrule(lr){2-10}
  & SuRe & 24.44 & 18.00 & 40.67 & 33.00 & 48.21 & 33.00 & 37.77 & 28.00 \\
  & Self-Ask  & 37.57 & \underline{28.50} & 50.53 & 39.50 & 45.12 & 35.00 & 44.41 & 34.33 \\
  & IRCoT & 29.83 & 20.50 & 46.36 & 36.50 & 48.79 & 36.50 & 41.66 & 31.17 \\
  & ItER-RETGEN & 36.53 & 26.50 & 55.16 & 45.50 & 58.63 & 44.50 & 50.11 & 38.83 \\
  & HippoRAG w/ IRCoT & 31.23 & 23.00 & 55.01 & 44.00 & 47.11 & 35.50 & 44.45 & 34.17 \\
  & LongRAG  & \underline{37.05} & 27.50 & \underline{60.49} & \underline{50.00} & \underline{62.64} & \underline{49.50} & \underline{53.39} & \underline{42.33} \\
  \cmidrule(lr){2-10}
  & \cellcolor{cyan!10}\approach
    & \cellcolor{cyan!10}\textbf{50.04} & \cellcolor{cyan!10}\textbf{39.00}
    & \cellcolor{cyan!10}\textbf{73.69} & \cellcolor{cyan!10}\textbf{64.00}
    & \cellcolor{cyan!10}\textbf{66.24} & \cellcolor{cyan!10}\textbf{51.00}
    & \cellcolor{cyan!10}\textbf{63.32} & \cellcolor{cyan!10}\textbf{51.33} \\
  \bottomrule
  \end{tabular}
  }
  \label{tab:main-results}
\end{table*}

\subsection{Answer Integration and Iterative Refinement}

Finally, the retrieved information is integrated to form the final answer. This step ensures that the hierarchical structure of the reasoning tree is respected, maintaining contextual relevance across sub-questions. If the initial retrieval fails to provide a satisfactory answer, the question is rephrased, and the decomposition and retrieval steps are repeated. This iterative process ensures robustness, allowing the framework to handle complex and ambiguous queries.

% \todo{check the placement of all the tables and figures}

\section{Experiments}
\subsection{Datasets}
We conduct experiments using three challenging multi-hop QA datasets widely recognized in the literature: MuSiQue~\citep{trivedi2022musique}, designed for evaluating multi-hop reasoning across diverse knowledge domains; 2WikiMQA~\citep{ho2020constructing}, which requires integrating information from two distinct Wikipedia articles; and HotpotQA~\citep{yang2018hotpotqa}, emphasizing complex question structures and paragraph-level evidence retrieval. Following previous studies~\citep{zhu2025mitigating,zhao2024longrag,bai2024longbench}, we adopt the same data splits and retrieval database configurations established in LongBench~\citep{bai2024longbench} to facilitate a direct comparison. The basic statistics of these datasets are presented in Table~\ref{tab:dataset_stats}.

\subsection{Baselines}
We compare \approach with several state-of-the-art multi-hop QA methods, which can be categorized as follows:
\begin{itemize}[leftmargin=0.5cm]
    \item \textbf{Direct}: This baseline involves directly prompting the LLM with the question without any retrieval augmentation, relying solely on the model's internal knowledge.
    \item \textbf{CoT}: Chain-of-Thought prompting~\citep{wei2023chainofthought} is used to encourage the LLM to generate a series of intermediate reasoning steps before providing the final answer.
    \item \textbf{NaiveRAG}: A standard retrieval-augmented generation approach where relevant documents are retrieved based on the original question, and the LLM generates an answer from the retrieved context.
    \item \textbf{NaiveRAG w/ QD}: An extension of NaiveRAG that first decomposes the complex question into simpler sub-questions (Query Decomposition) and then retrieves documents for each sub-question.
    \item \textbf{SuRe}: A method that involves generating a scene and then retrieving relevant information based on that scene to answer the question~\citep{kim2023sure}.
    \item \textbf{IRCoT}: Iterative Retrieval with Chain-of-Thought, which combines iterative retrieval with CoT prompting to refine the reasoning process over multiple steps~\citep{trivedi2023interleaving}.
    \item \textbf{Self-Ask}: An iterative approach where the LLM explicitly asks and answers follow-up questions to gather intermediate facts before answering the main question~\citep{press2023measuring}.
    \item \textbf{ItER-RETGEN}: An iterative retrieval and generation framework that refines its answers over multiple iterations~\citep{shao2023enhancing}.
    \item \textbf{HippoRAG}: A retrieval framework inspired by the hippocampal indexing theory of human long-term memory to enable deeper and more efficient knowledge integration over new experiences.~\citep{gutierrez2024hipporag}.
    \item \textbf{LongRAG}: A RAG method optimized for long contexts, which also uses an iterative retrieval process to handle extensive documents~\citep{zhao2024longrag}.
    \item \textbf{ChainRAG}: A graph-based structured decomposition approach for multi-hop QA that models the reasoning process as a chain or graph of retrieval and reasoning steps~\citep{zhu2025mitigating}.
\end{itemize}

\subsection{Metrics}
We evaluate model performance using standard metrics prevalent in multi-hop QA research~\citep{zhu2025mitigating,zhao2024longrag,trivedi2023interleaving}: Exact Match (EM) and F1 scores. EM assesses whether the predicted answer exactly matches one of the ground truth answers, while F1 evaluates token-level overlap, capturing partial correctness through precision and recall.

\subsection{Experimental Setup}

To ensure a fair and consistent evaluation settings across all methods, we standardize the key experimental components. We utilize \texttt{text-embedding-3-small}\footnote{\url{https://platform.openai.com/docs/guides/embeddings}} as the embedding model and the \texttt{bge-reranker-base}\footnote{\url{https://huggingface.co/BAAI/bge-reranker-base}} as the reranker, aligning with the configuration in ChainRAG~\citep{zhu2025mitigating}. Our evaluations leverage both open-source and closed-source LLMs, specifically \texttt{Qwen-2.5-14B-Instruct}~\citep{qwen2025qwen25} and \texttt{GPT-4o-mini}\footnote{\url{https://platform.openai.com/docs/models/gpt-4o-mini}}. 

\begin{table*}[t]
    \centering
    % \resizebox{\textwidth}{!}{
    \caption{Ablation study results on MusiQue, 2WiKiMQA, and HotpotQA using Qwen2.5-14B-Instruct model.}
    \vspace{-0.25cm}
    \begin{tabular}{lcccccc} 
    \toprule
    \multirow{2}{*}{Configuration} & \multicolumn{2}{c}{MusiQue} & \multicolumn{2}{c}{2WiKiMQA} & \multicolumn{2}{c}{HotpotQA} \\
    \cmidrule(lr){2-3} \cmidrule(lr){4-5} \cmidrule(lr){6-7}
     & F1 & EM & F1 & EM & F1 & EM \\
    \midrule
\approach & 50.04 & 39.00 & 73.69 & 64.00 & 66.24 & 51.00 \\
\quad w/o Consensus-Based Tree Selection & 49.27 {\scriptsize (\textcolor{red!40}{-0.77})} & 37.50 {\scriptsize (\textcolor{red!40}{-1.50})} & 72.03 {\scriptsize (\textcolor{red!40}{-1.66})} & 61.00 {\scriptsize (\textcolor{red!100}{-3.00})} & 63.16 {\scriptsize (\textcolor{red!100}{-3.08})} & 50.00 {\scriptsize (\textcolor{red!40}{-1.00})} \\
\quad w/o Rejection Sampling (Retrieval) & 47.97 {\scriptsize (\textcolor{red!80}{-2.07})} & 37.00 {\scriptsize (\textcolor{red!80}{-2.00})} & 72.95 {\scriptsize (\textcolor{red!40}{-0.74})} & 62.00 {\scriptsize (\textcolor{red!80}{-2.00})} & 63.89 {\scriptsize (\textcolor{red!80}{-2.35})} & 49.00 {\scriptsize (\textcolor{red!80}{-2.00})} \\
\quad w/o Query Rewriting & 47.09 {\scriptsize (\textcolor{red!80}{-2.95})} & 36.50 {\scriptsize (\textcolor{red!80}{-2.50})} & 71.42 {\scriptsize (\textcolor{red!80}{-2.27})} & 61.00 {\scriptsize (\textcolor{red!100}{-3.00})} & 65.08 {\scriptsize (\textcolor{red!40}{-1.16})} & 51.00 {\scriptsize (\textcolor{black}{+0.00})} \\
\quad w/o Structural Analysis & 48.82 {\scriptsize (\textcolor{red!40}{-1.22})} & 37.00 {\scriptsize (\textcolor{red!80}{-2.00})} & 72.74 {\scriptsize (\textcolor{red!40}{-0.95})} & 63.00 {\scriptsize (\textcolor{red!40}{-1.00})} & 63.58 {\scriptsize (\textcolor{red!80}{-2.66})} & 49.00 {\scriptsize (\textcolor{red!80}{-2.00})} \\
    \bottomrule 
    \end{tabular} 
    % }
    % Performance change compared to the full model is shown in parentheses.}
    \label{tab:ablation}  
\end{table*}

Our overall experimental design follows the setup detailed in ChainRAG~\citep{zhu2025mitigating}. For baseline methods requiring re-implementation on our part, document processing typically utilized a chunk size of 200 tokens with a 100-token overlap. Retrieval parameters were consistently set to $k=45$ for coarse ranking, $k=15$ for fine ranking, and a 3000-token limit for retrieved context, unless a baseline's original methodology specified different configurations. Comparative baseline results were either adopted from their original publications, sourced from ChainRAG~\citep{zhu2025mitigating} for specific model configurations as detailed below, or obtained through our re-implementations based on publicly available code or official algorithmic descriptions. As for other hyperparameters in \approach, the number of candidates in the consensus-based tree selection and the rejection sampling in answer generation are set to 5, the rounds of iterative refinement was limited up to 3. For experiments utilizing the GPT-4o-mini model, results for NaiveRAG, NaiveRAG with Query Decomposition (QD), ITER-RETGEN~\citep{shao2023enhancing}, LongRAG~\citep{zhao2024longrag}, HippoRAG with IRCoT~\citep{gutierrez2024hipporag}, and ChainRAG itself were directly adopted from~\citet{zhu2025mitigating}. All other baseline results, particularly those involving the Qwen-2.5-14B model, were obtained through our re-implementations. Specifically: IRCoT was re-implemented based on its original publication, incorporating details from its description in the HippoRAG paper~\citep{gutierrez2024hipporag}. SelfAsk was re-implemented based on its original publication, also referencing its description within the EfficientRAG paper~\citep{zhuang2024efficientrag}. For ItER-RETGEN~\citep{yue2024inference}, in the absence of publicly available code, we developed an implementation based on its published prompt structures. LongRAG was implemented using its official open-source repository and default settings, which include word-count based document segmentation with a chunk size of 200. HippoRAG~\citep{gutierrez2024hipporag} was also implemented, adhering to its design which omits document chunking. Due to the unavailability of its source code, reproducing ChainRAG results with the Qwen-2.5-14B model was infeasible. Consequently, for ChainRAG, comparisons are made against its GPT-4o-mini performance as reported in the original paper~\citep{zhu2025mitigating}.

% Full Model: F1 avg = 63.32, EM avg = 51.33  
% w/o Consensus-Based Tree Selection: F1 avg = 61.49, EM avg = 49.50  
% w/o Rejection Sampling (Retrieval): F1 avg = 61.60, EM avg = 49.33  
% w/o Query Rewriting: F1 avg = 61.20, EM avg = 49.50  
% w/o Structural Analysis: F1 avg = 61.71, EM avg = 49.67  

\subsection{Results and Analysis}

Table~\ref{tab:main-results} presents a comprehensive evaluation of \approach against existing multi-hop QA methods across three benchmarks.

\lparagraph{Overall Performance} \approach achieves state-of-the-art results across all evaluated benchmarks, demonstrating substantial improvements. For instance, with the GPT-4o-mini model, \approach attains an average F1 score 7.0\% higher and an EM score 6.0\% higher than the leading ChainRAG variant. This advantage expands with the Qwen2.5-14B-Instruct model, where \approach surpasses LongRAG by 9.9\% in F1 and 9.0\% in EM. These results underscore the architecture-agnostic efficacy of \approach. Its core distinction lies in its methodology: \approach first performs explicit problem structure analysis and then establishes a consensus-driven tree to manage sub-questions before retrieval. This structured pre-retrieval planning creates clear reasoning pathways, offering a more cohesive global reasoning framework compared to techniques that integrate reasoning and retrieval iteratively (e.g., IRCoT, ItER-RETGEN, Self-Ask). Consequently, \approach also demonstrates superior results over other structured methods like ChainRAG and LongRAG. This structured approach, coupled with its consensus mechanism, is key to enhancing the robustness of reasoning and retrieval operations, thereby effectively minimizing error propagation and hallucination in complex multi-hop QA tasks.

\lparagraph{Dataset-Specific Analysis} \approach exhibits particularly strong performance on 2WikiMQA, improving the F1 score by 12.5\% and the EM score by 11.0\% with GPT-4o-mini, and achieving gains of 13.2\% in F1 and 14.0\% in EM with Qwen2.5-14B. This substantial improvement aligns with 2WikiMQA's unique construction combining structured Wikidata triples with unstructured Wikipedia text, which creates reasoning paths requiring precise decomposition and evidence tracking—characteristics well-matched to our tree-guided approach. On MuSiQue, we observe significant improvements of 3.9\% in F1 and 3.0\% in EM with GPT-4o-mini, and 13.0\% in F1 and 11.5\% in EM with Qwen2.5-14B. These gains reflect MuSiQue's distinctive bottom-up design that enforces connected reasoning through carefully composed single-hop questions across 2-4 hops and six different reasoning structures, creating a challenging but structurally decomposable reasoning environment. Although the gains on HotpotQA are more modest (0.7\% and 3.6\% F1, 1.5\% EM for both models), they remain notable considering HotpotQA's known susceptibility to reasoning shortcuts and single-hop solutions that limit the effectiveness of explicit decomposition strategies.

\lparagraph{Model-Specific Performance} Our findings indicate that \approach effectively enhances the capabilities of different model architectures. While GPT-4o-mini serves as a strong baseline, results for Qwen2.5-14B-Instruct with \approach frequently match or exceed those of GPT-4o-mini across several metrics. This suggests that the clear reasoning pathways established by \approach's explicit problem structure analysis and consensus-driven tree are particularly beneficial, aiding models in navigating complex multi-hop queries, especially those with varied intrinsic reasoning capabilities.

These empirical findings validate our theoretical framework: modeling a hierarchical reasoning structure through trees with consensus-based selection and bottom-up synthesis effectively addresses core challenges in multi-hop QA, particularly error propagation and hallucination.

\begin{table*}[t]
\renewcommand{\arraystretch}{1.2} 
\centering
\footnotesize
\caption{Comparison of reasoning chains: The left column shows that Self-Ask mistakenly treats Felix Salten’s \textbf{\textcolor{blue}{birthplace}} as his \textbf{\textcolor{blue}{home city}}. The right column demonstrates our method, which correctly identifies \textcolor{green}{Vienna} as the home city by following a sound reasoning path. Irrelevant steps are highlighted in \textcolor{orange}{orange}.}
\begin{tabular}{|p{0.48\textwidth}|p{0.48\textwidth}|}
\hline
\textbf{Reasoning Chain of Self-Ask} & \textbf{Reasoning Chain of RT-RAG} \\
\hline
\multicolumn{2}{|p{0.96\textwidth}|}{
\textbf{Question:} What was the \textbf{\textcolor{blue}{home city}} of the author of the famous novel tracing the life of a male roe deer from his birth onward? \newline
\textbf{Correct Answer:} \textcolor{green}{Vienna}
} \\
\hline

\textbf{Self-Ask} \par
Are follow up questions needed here: \textbf{Yes} \par
\textbf{Follow up:} Who is the author of the famous novel tracing the life of a male roe deer from his birth onward? \par
\textbf{Intermediate answer:} \textbf{\textcolor{blue}{Felix Salten}} \par
\textcolor{orange}{\textbf{Follow up:} What is the title of the famous novel by Felix Salten?} \par
\textcolor{orange}{\textbf{Intermediate answer:} \textit{Bambi: Eine Lebensgeschichte aus dem Walde}} \par
\textbf{Follow up:} Where was Felix Salten \textbf{\textcolor{red}{born}}? \par
\textbf{Intermediate answer:} \textcolor{red}{Pest, Austria-Hungary} \par
\textbf{The final answer is:} \textcolor{red}{\textbf{Pest, Austria-Hungary}}\par
\vspace{2.25em}
\rule{\linewidth}{0.4pt} \par
\textbf{Final Answer:} \textcolor{red}{\textbf{Pest, Austria-Hungary}} 
&

\textbf{Question decomposition} \par
\textbf{Left child (N1):} Who is the author of the famous novel tracing the life of a male roe deer from his birth onward? \par
\textbf{Right child (N2):} What was the \textbf{\textcolor{blue}{home city}} of [answer from N1] \par
\rule{\linewidth}{0.4pt} \par
\textbf{Reasoning Steps:} \par
\textbf{Step 1:} Attempt to decompose N1 → leaf node \par
\textbf{Step 2:} Retrieve N1 → \textbf{\textcolor{blue}{Felix Salten}} \par
\textbf{Step 3:} Substitute N2: What was the \textbf{\textcolor{blue}{home city}} of \textbf{Felix Salten} \par
\textbf{Step 4:} Attempt to decompose N2 → leaf node \par
\textbf{Step 5:} Retrieve N2 → \textbf{\textcolor{blue}{Vienna}} \par
\textbf{Step 6:} Aggregate N1 and N2 → \textcolor{green}{Vienna} \par
\rule{\linewidth}{0.4pt} \par
\textbf{Final Answer:} \textcolor{green}{\textbf{Vienna}} \\
\hline
\end{tabular}
\label{tab:case}
\end{table*}

\begin{table}[h]
    \centering
    \caption{Impact of max tree depth on performance across different datasets with Qwen2.5-14B-Instruct model.}
    \begin{adjustbox}{width=\columnwidth}
    \begin{tabular}{cccccccc}
    \toprule
    \multirow{2}{*}{Max Depth} & \multicolumn{2}{c}{MusiQue} & \multicolumn{2}{c}{2WikiMQA} & \multicolumn{2}{c}{HotpotQA} \\
    \cmidrule(lr){2-3} \cmidrule(lr){4-5} \cmidrule(lr){6-7}
    & F1 & EM & F1 & EM & F1 & EM \\
    \midrule
    1 & 38.70 & 28.50 & 57.95 & 46.50 & 60.24 & 45.00 \\
    2 & 47.63 & 37.00 & 73.61 & 62.50 & 64.98 & 49.50 \\
    3 & 49.57 & 38.50 & \textbf{73.80} & 63.00 & 65.80 & 50.50 \\
    4 & \textbf{50.04} & \textbf{39.00} & 73.69 &\textbf{64.00} & \textbf{66.24} & \textbf{51.00} \\
    \bottomrule
    \end{tabular}
    \end{adjustbox}
    \label{tab:tree_depth}
\end{table}

\subsection{Ablation Study}
To assess the individual contributions of key components within the \approach framework, we performed a comprehensive ablation study using the Qwen2.5-14B-Instruct model, with results detailed in Table~\ref{tab:ablation}.
On average, removing \textit{Query Rewriting} incurred the most significant F1 score reduction of 2.1\%, while disabling \textit{Rejection Sampling} led to the largest EM score decrease of 2.0\%. \textit{Consensus-Based Tree Selection} also proved highly influential; its absence resulted in an average decrease of 1.8\% for both F1 and EM scores. \textit{Structural Analysis}, though its removal yielded the smallest average performance decline, still demonstrated its importance, as evidenced by F1 and EM score reductions of 1.6\% and 1.7\%, respectively. These average figures highlight the critical role of each component.

Beyond these aggregate effects, dataset-specific analyses reveal further insights. For instance, the utility of \textit{Query Rewriting} is particularly evident for datasets featuring complex or nuanced queries, such as MuSiQue and 2WikiMQA, where its removal led to F1 score reductions of 3.0\% and EM score reductions of 3.0\%, respectively. Conversely, its omission had a negligible impact on the EM score for HotpotQA, suggesting that HotpotQA queries might be more direct or less prone to ambiguity affecting exact matches. Another salient observation is the pronounced F1 score degradation on HotpotQA when \textit{Consensus-Based Tree Selection} or \textit{Structural Analysis} was excluded, leading to F1 score decreases of 3.1\% and 2.7\%, respectively. This underscores the critical need for robust reasoning path selection and explicit problem decomposition for HotpotQA, a dataset characterized by questions often requiring comparative reasoning and multi-step information synthesis.
These findings collectively affirm that each ablated component contributes significantly to the overall effectiveness and robustness of the \approach framework.

\begin{table*}[htbp]
\centering
\caption{Hierarchical Question Decomposition and Reasoning Process for a Multi-hop Question about the Italian Navigator with \approach. A key strength shown here is the ability to handle missing intermediate answers \textcolor{red}{\textbf{[none]}} by dynamically converting the parent node into a new leaf, ensuring continuity in reasoning and preserving answer accuracy in multi-hop settings.}
\setlength{\parskip}{0pt}
\scriptsize

\begin{tcolorbox}[width=\textwidth, colback=white, boxsep=4pt, left=4pt, right=4pt, top=4pt, bottom=4pt]
\textbf{Original Question} \\
Who is the son of the Italian navigator who explored the eastern coast of the continent Ulises Solís' birthplace is located in for England?\\
\textbf{Correct Answer:Sebastian Cabot}\\
\vspace{1.5pt}
\noindent\rule{\textwidth}{0.4pt}

\textbf{Initial Decomposition} \\
\textbf{Question:} Who is the son of the Italian navigator who explored the eastern coast of the continent Ulises Solís' birthplace is located in for England? \\
\hspace*{1em}\textbf{Left child: N1} — Who is the Italian navigator who explored the eastern coast of the continent where Ulises Solís' birthplace is located? \\
\hspace*{2em}\textbf{Left child: N2} — In which continent was Ulises Solís born? \\
\hspace*{3em}\textbf{Left child: N3} — Where was Ulises Solís born? \\
\hspace*{3em}\textbf{Right child: N4} — In which continent is [answer from N3] located? \\
\hspace*{2em}\textbf{Right child: N5} — Who is the Italian navigator who explored the eastern coast of [answer from N2]? \\
\hspace*{1em}\textbf{Right child: N6} — Who is the son of [answer from N1]?

\vspace{1.5pt}
\noindent\rule{\textwidth}{0.4pt}

\textbf{Non-leaf to Leaf Node Conversion} \\
Step 1: Retrieve answer for N3— Q: Where was Ulises Solís born? A: \textcolor{red}{\textbf{[none]}} \\
Step 2: As N3's answer is \textcolor{red}{\textbf{[none]}}, we update its parent node N2 to be a new leaf node.

\vspace{1.5pt}
\noindent\rule{\textwidth}{0.4pt}

\textbf{Final Decomposition} \\
\textbf{Question:} Who is the son of the Italian navigator who explored the eastern coast of the continent Ulises Solís' birthplace is located in for England? \\
\hspace*{1em}\textbf{Left child: N1} — Who is the Italian navigator who explored the eastern coast of the continent where Ulises Solís' birthplace is located? \\
\hspace*{2em}\textbf{Left child: N2} — In which continent was Ulises Solís born? \\
\hspace*{2em}\textbf{Right child: N5} — Who is the Italian navigator who explored the eastern coast of [answer from N2]? \\
\hspace*{1em}\textbf{Right child: N6} — Who is the son of [answer from N1]?

\vspace{1.5pt}
\noindent\rule{\textwidth}{0.4pt}

\textbf{Reasoning Step} \\
Step 1: Retrieve answer for N2 → \textcolor{blue}{\textbf{North America}} \\
Step 2: N5 becomes — Who is the Italian navigator who explored the eastern coast of \textcolor{blue}{\textbf{North America}}? \\
Step 3: Attempt to decompose the new N5 question→Determine N5 as a leaf node \\
Step 4: Retrieve answer for N5 → \textcolor{blue}{\textbf{John Cabot}} \\
Step 5: Aggregate N2 and N5 to answer N1  → \textcolor{blue}{\textbf{John Cabot}} \\
Step 6: N6 becomes — Who is the son of \textcolor{blue}{\textbf{John Cabot}}? \\
Step 7: Attempt to decompose the new N6 question→Determine N6 as a leaf node \\
Step 8: Retrieve answer for N6 → \textcolor{blue}{\textbf{Sebastian Cabot}} \\
Step 9: Aggregate N1 and N6 to answer original question →   \textcolor{blue}{\textbf{Sebastian Cabot}}

\vspace{1.5pt}
\noindent\rule{\textwidth}{0.4pt}

\textbf{Final Answer} \\
\textcolor{blue}{\textbf{Sebastian Cabot}}
\end{tcolorbox}

\label{tab:cabot-decomposition}
\end{table*}

\subsection{Impact of Tree Depth}
We examined the relationship between tree depth constraints and model performance by systematically varying the maximum depth of reasoning trees, with results presented in Table~\ref{tab:tree_depth}. Performance demonstrates a clear progression with increasing depth. Advancing from depth 1 to 2 results in substantial improvements across all datasets, particularly on 2WikiMQA, where the F1 score increased by 15.7\% and the EM score by 16.0\%. The transition from depth 2 to 3 yields more modest gains, with MuSiQue showing the most notable improvement, where its F1 score increased by 1.9\% and its EM score by 1.5\%. Further increasing the depth to 4 leads to minimal performance changes, suggesting that depth 3 represents an optimal balance between decomposition granularity and computational efficiency.

\begin{figure}[ht]
\centering
\includegraphics[width=0.8\columnwidth]{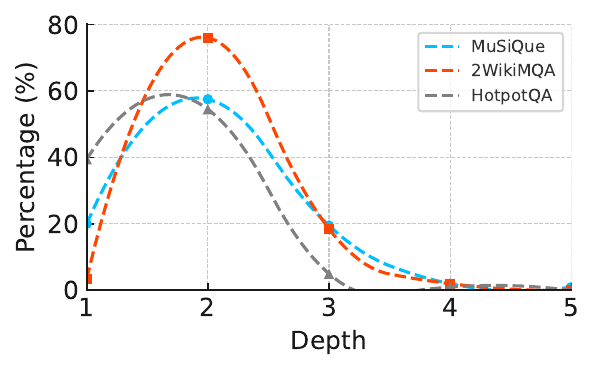}
\vspace{-10pt}
\caption{Tree depth distribution on different datasets with Qwen2.5-14B-Instruct model.}
\label{fig:tree-depth}
\end{figure}

Figure~\ref{fig:tree-depth} illustrates the distribution of actual tree depths across datasets under a maximum depth constraint of 5. MuSiQue and 2WikiMQA predominantly leverage depth-2 trees (72\% and 76\%, respectively), indicating that moderate complexity suffices for their reasoning needs. In contrast, HotpotQA exhibits 39\% of questions effectively addressed by depth-1 trees, reflecting its simpler question structures. The prevalence of depth-2 trees in MuSiQue and 2WikiMQA aligns with their typical reasoning patterns: 2WikiMQA’s strictly two-hop questions and MuSiQue’s common 2–4 hop chains. MuSiQue also features some depth-3 and depth-4 trees, allowing for more complex reasoning. In comparison, HotpotQA’s larger share of depth-1 trees highlights its lower multi-hop demand. These distributions reflect the datasets’ inherent complexity profiles and guide optimal configuration for the \approach framework.

% Figure~\ref{fig:tree-depth} illustrates the distribution of actual tree depths across datasets under a maximum depth constraint of 5. MuSiQue and 2WikiMQA predominantly leverage depth-2 trees (72\% and 76\%, respectively), demonstrating that moderate complexity effectively meets their reasoning demands. In contrast, HotpotQA showcases a broader distribution, with 39\% of questions most effectively addressed by depth-1 trees, reflecting its relatively simpler question structures.

% The prevalence of depth-2 trees in MuSiQue and 2WikiMQA aligns closely with 2WikiMQA’s strictly two-hop reasoning and MuSiQue’s typical chains involving 2–4 hops, where many questions naturally decompose into two main reasoning steps. Additionally, MuSiQue’s modest representation of depth-3 and depth-4 trees allows for longer multi-step inferences, accommodating more complex reasoning scenarios. Conversely, HotpotQA’s simpler structure, characterized by many questions answerable in a single step, explains why over a third of its instances optimally utilize depth-1 trees, underscoring its lower multi-hop demand compared to the other benchmarks. 
% These distributions align with the inherent complexity profiles of each dataset and inform optimal configuration parameters for the \approach framework.

% This analysis highlights the varying complexities of reasoning required across different datasets, reflecting their distinct design philosophies and question structures.

\section{Case Study}
\label{sec:case-study}

Table~\ref{tab:case} compares \approach with Self-Ask using an example from HotPotQA, illustrating how structured decomposition yields more reliable reasoning than self-guided exploration.
Both methods initially identify Felix Salten correctly, but Self-Ask drifts by retrieving irrelevant book titles and confusing his birthplace with his home city, resulting in an incorrect answer.
In contrast, \approach maintains focus through its hierarchical structure: by decomposing the query into relevant, dependency-linked sub-questions, it ensures each retrieval targets the needed information and correctly identifies Vienna as Salten’s home city.
This case highlights how \approach’s structured reasoning naturally constrains hallucination and prevents error propagation.

Furthermore, the example in Table~\ref{tab:cabot-decomposition} highlights the robustness of \approach in handling complex queries with unanswerable sub-questions. The query depends on identifying the birthplace of “Ulises Solís,” which is absent from the knowledge source. While most methods fail when encountering such gaps, \approach dynamically restructures the reasoning path: when the retrieval for node N3 returns no answer, its parent node N2 is converted into a new leaf node. This adjustment enables the model to bypass the missing information, seek the continent directly, and ultimately derive the correct answer—North America. This case underscores a major strength of our hierarchical framework: its ability to gracefully recover from retrieval failures through dynamic reconfiguration, maintaining robustness in multi-hop reasoning with incomplete information.

Together, these two case studies illustrate the dual strengths of \approach: precise, hallucination-resistant reasoning through structured decomposition, and resilience through adaptive restructuring. By combining logical focus with flexibility, \approach offers a robust and reliable solution for multi-hop question answering.

\section{Conclusion}
In this paper, we introduced Reasoning Tree Guided RAG, a novel hierarchical framework for multi-hop question answering. \approach transforms complex questions into structured tree decompositions via a consensus-based selection mechanism, ensuring robust and clear reasoning pathways. Key innovations, including adaptive leaf node determination to prevent over-decomposition and rejection sampling during retrieval to minimize hallucinations, contribute to its efficacy. Our experiments demonstrate that \approach significantly outperforms existing methods on established multi-hop QA benchmarks, achieving state-of-the-art results. By explicitly modeling the hierarchical nature of reasoning and integrating structured decomposition with dynamic retrieval and answer aggregation, \approach offers a promising direction for enhancing the reliability and accuracy of RAG systems in complex information-seeking tasks.

\clearpage
% \section*{Limitations} 

% While \approach demonstrates strong performance in multi-hop QA, several avenues warrant further exploration. First, adapting our structured reasoning and tree decomposition paradigm effectively to a broader range of complex generative tasks beyond question answering, such as argumentative text generation or multifaceted summarization, requires further investigation into defining appropriate reasoning nodes for diverse output modalities. Second, deeper integration of the tree generation and traversal process with LLM fine-tuning could further enhance robustness, potentially training models to explicitly generate optimal decompositions and refine reasoning within the tree structure, thereby better mitigating errors and hallucination. Finally, applying \approach to highly specialized domains may necessitate adaptations to our current problem structure analysis and consensus mechanisms to ensure optimal performance when encountering unique terminologies or implicit relational structures not prevalent in general corpora.

% \section*{Acknowledgments}

\newpage
\bibliographystyle{ACM-Reference-Format}
\bibliography{reference}

% \clearpage
% \appendix

% \input{appendix.tex}

\end{document}